\documentclass[conference]{IEEEtran}
\IEEEoverridecommandlockouts
\usepackage{cite}
\usepackage{graphicx}
\usepackage{color,array}
\usepackage{csquotes}
\usepackage{graphicx}
\usepackage{lineno,hyperref}
\usepackage{amsmath,amsthm,amssymb,amsfonts}
\usepackage[ruled,vlined]{algorithm2e}
\usepackage{textcomp}
\usepackage{float}
\usepackage{diagbox}
\usepackage{caption}
\usepackage{subcaption}
\usepackage{tabularx}    
\usepackage{multirow}
\usepackage{algpseudocode}
\usepackage{xcolor}
\usepackage{adjustbox}
\usepackage{booktabs} 
\usepackage{siunitx}
\usepackage{hhline}
\usepackage{orcidlink}

\def\BibTeX{{\rm B\kern-.05em{\sc i\kern-.025em b}\kern-.08em
    T\kern-.1667em\lower.7ex\hbox{E}\kern-.125emX}}
    
\begin{document}

\title{Exploiting Meta-Learning-based Poisoning Attacks for Graph Link Prediction}

\author{
\IEEEauthorblockN{Mingchen Li\footnotemark*}
\IEEEauthorblockA{\textit{University of South Florida} \\
Tampa, United States \\}
\and
\IEEEauthorblockN{Di Zhuang}
\IEEEauthorblockA{\textit{Snap Inc.} \\
Santa Monica, United States \\}
\and
\IEEEauthorblockN{Keyu Chen}
\IEEEauthorblockA{\textit{Snap Inc.} \\
Santa Monica, United States \\}
\and
\IEEEauthorblockA{\vspace{-6.5ex}} 
\and
\IEEEauthorblockN{\hspace{2.5cm} Dumindu Samaraweera} 
\IEEEauthorblockA{\hspace{2.5cm} \textit{Embry-Riddle Aeronautical University} \\ 
\hspace{2.5cm} Daytona Beach, United States\\}
\and
\IEEEauthorblockN{\hspace{-1.5cm}J. Morris Chang}
\IEEEauthorblockA{\hspace{-1.5cm}\textit{University of South Florida} \\
\hspace{-1.5cm}Tampa, United States \\}
}

\maketitle
\thispagestyle{plain} 
\pagestyle{plain}    

\begin{abstract}
Link prediction in graph data uses various algorithms and Graph Nerual Network (GNN) models to predict potential relationships between graph nodes. These techniques have found widespread use in numerous real-world applications, including recommendation systems, community/social networks, and biological structures. However, recent research has highlighted the vulnerability of GNN models to adversarial attacks, such as poisoning and evasion attacks. Addressing the vulnerability of GNN models is crucial to ensure stable and robust performance in GNN applications. Although many works have focused on enhancing the robustness of node classification on GNN models, the robustness of link prediction has received less attention. To bridge this gap, this article introduces an unweighted graph poisoning attack that leverages meta-learning with weighted scheme strategies to degrade the link prediction performance of GNNs. We conducted comprehensive experiments on diverse datasets across multiple link prediction applications to evaluate the proposed method and its parameters, comparing it with existing approaches under similar conditions. Our results demonstrate that our approach significantly reduces link prediction performance and consistently outperforms other state-of-the-art baselines.
\end{abstract}


\renewcommand{\thefootnote}{\fnsymbol{footnote}}
\footnotetext[1]{Mingchen Li and J. Morris Chang (mingchenli@usf.edu, chang5@usf.edu) are affiliated with the Department of Electrical Engineering at the University of South Florida. Di Zhuang and Keyu Chen (zhuangdi1990@gmail.com, keyuchen07@gmail.com) were previously associated with the University of South Florida, and are currently with Snap Inc.. Their contributions to this work were made during his tenure at the University of South Florida. Dumindu Samaraweera (samarawg@erau.edu) is currently affiliated with Embry-Riddle Aeronautical University}

\section{Introduction}
Many real-world datasets such as social networks~\cite{borgatti2009network}, traffic networks~\cite{latora2002boston}, biological structures~\cite{montoya2002small}, and e-commerce networks~\cite{zeng2019graphsaint} can be represented as graphs containing nodes, links, and associated features. The structure of these graphs is dynamic, constantly changing as relationships evolve. Capturing these changes is fundamental to the success of various real-world scenarios, such as community detection and recommendation systems~\cite{chen2017improved}. Node classification and link prediction are the primary tasks that address this challenge, employing various algorithms and graph neural network models to predict potential labels (nodes) and relationships (links) between nodes in a graph.

During the past few decades, numerous graph approaches have emerged in various research domains, including Local Random Walk~\cite{liu2010link}, DeepWalk~\cite{perozzi2014deepwalk}, node2vec~\cite{grover2016node2vec}, Graph Convolutional Networks (GCN)~\cite{kipf2016semi}, Variational Graph Autoencoder (VGAE)~\cite{kipf2016variational}, Graph Attention Network (GAT)~\cite{velivckovic2017graph} and others. However, recent studies have underscored the susceptibility of graph learning models to adversarial attacks~\cite{jin2021adversarial}, such as poisoning and evasion attacks, which pose significant threats to the performance and stability of graph models. To exploit these vulnerabilities, numerous approaches of graph adversarial attacks have been proposed. Xu et al.~\cite{xu2019topology} proposed a topology attack with projected gradient descent (PGD) from an optimization perspective to optimize the negative cross-entropy using the gradient to maximize the prediction error of the model. Waniek et al.~\cite{waniek2018hiding} proposed the Disconnect Internally, Connect Externally (DICE) algorithm, which reduces the internal graph density to deceive graph models. Chen et al.~\cite{chen2018link} used an adversarial network generator to initiate gradient-based attacks, effectively misleading state-of-the-art link prediction algorithms. Furthermore, Meta-learning~\cite{finn2017model}, offering valuable insights for efficient learning and rapid adaptation, has emerged as an attack strategy. In~\cite{zugner_adversarial_2019}, the authors utilized meta-learning to conduct adversarial modifications on graph data. A key observation is that most existing methods primarily focus on node classification tasks, while the adversarial vulnerability of link prediction remains comparatively underexplored. This represents a significant and important gap in the current research landscape.


In this study, we aim to bridge this gap by investigating adversarial attacks on graph link prediction and propose an unweighted graph poisoning attack approach. Leveraging meta-learning techniques with weighted schemes during the adversarial training stage, our method aims to degrade the performance of link prediction models. Through extensive experiments on diverse datasets and corresponding use cases, we evaluated the effectiveness of our approach and the impact of the parameters. In addition, we compare our method with existing approaches, demonstrating its superior effectiveness. To ensure reproducibility and facilitate further research, we have made the source code publicly available at our github repository\footnote[1]{\href{https://github.com/mingchenli/VGAE_Attack_Meta}{https://github.com/mingchenli/VGAE\_Attack\_Meta}}.

The main contributions of this work are as follows:
\begin{itemize}
  \item We propose a novel poisoning attack that leverages meta-learning with weighted schemes to effectively degrade link-prediction performance.
  \item We introduce three weighted schemes within the meta-learning framework to identify optimal attack directions, enabling subtle yet impactful adversarial graph modifications.
  \item We present a comparative analysis of these weighted schemes, experimentally validating our design and providing key insights for the future development of weighted strategies.  
  \item We conduct extensive experiments on benchmark datasets across diverse use cases, analyzing parameter effects and demonstrating that our method outperforms state-of-the-art approaches.
\end{itemize}

The remaining sections of the paper are structured as follows: Section \ref{sec:Related work} provides an overview of related works, Section \ref{sec:Background} offers background knowledge on our approach, Section \ref{sec:Methodology} outlines our methodology, Section \ref{sec: Experimental Evaluation} presents the experimental evaluation, and Section \ref{sec:Conclusion} concludes the paper.

\section{Related Works}\label{sec:Related work}

Adversarial attacks on graph data have gained significant traction as a crucial topic in machine learning research in recent years. These attacks serve to identify model weaknesses and bolster model robustness through the development of corresponding defense mechanisms. Even minor and unnoticeable adjustments to graph data can lead to notable changes in model performance, underscoring the importance of addressing vulnerabilities in these machine learning models. Over the past few years, several effective attack approaches have been devised specifically for graph data and models. 

\subsection{Gradient-Based Attacks}
Gradient-based adversarial attacks utilize model gradients to identify and perturb critical graph elements. Xu et al.~\cite{xu2019topology} proposed projected gradient descent (PGD) and MinMax attacks that treat the attack as an optimization problem. They have proposed an optimization framework to optimize the negative cross-entropy using the gradient to maximize the prediction error of the model. Their method uses PGD to iteratively modify links, significantly affecting model performance while respecting a budget constraint on the number of modifications. Chen et al.~\cite{chen2018link} introduced the iterative gradient attack (IGA) for the attack on link prediction tasks. This method leverages gradients derived from trained graph auto-encoders to identify and perturb the significant links in the graph. They perform comprehensive evaluations to demonstrate that even minor link alterations can disrupt the performance of state-of-the-art link prediction models. Zhang et al.~\cite{zhang2022unsupervised} developed a gradient-based attack based on graph contrastive learning. By maximizing the contrastive loss during the training phase, the attack disrupts the embeddings produced by the contrastive model under an unsupervised learning setting, making it suitable for real-world where sometimes the data labels are scarce. In~\cite{zugner_adversarial_2019}, a meta-learning based poisoning attack strategy was proposed to target the significant part of the graph structure or node characteristics. They used meta-gradient as guidance to generate adversarial modifications to decrease graph model performance. 

\subsection{Heuristic-Based Attacks}
Heuristic-based attacks employ predefined strategies to perturb graphs, often drawing on principles from graph theory. Zügner et al. proposed Nettack~\cite{zugner2018adversarial}, a framework that applies a greedy algorithm to identify and modify critical links and node attributes, thereby inducing misclassifications in node-classification tasks. Their results show that model vulnerabilities are strongly tied to graph structure in semi-supervised settings.The DICE attack introduced in~\cite{waniek2018hiding} follows the Disconnect Internally, Connect Externally (DICE) strategy to manipulate community density and disrupt specific targets. By strategically altering the graph structure, DICE effectively degrades the performance of graph-based models.

\subsection{Reinforcement Learning-Based Attacks} 
Reinforcement learning-based methods utilize a sequential decision-making process with the agent to achieve the adversarial attack by using a reward system to maximize misclassification rates in black-box settings. Dai et al.~\cite{dai2018adversarial} presented a reinforcement learning framework that modifies graph structures through link additions or deletions. Sun et al.~\cite{sun2019node} extended reinforcement learning to node injection attacks, where fake nodes are introduced into the graph. A reinforcement learning agent is used to determine the optimal connections for these nodes, maximizing the overall classification error while keeping the original structure of the graph similar.

However, most of the attacks described above are aimed at graph convolutional networks~\cite{kipf2016semi} and node classification. GCNs face limitations in representational capacity, supported graph types, and generative abilities. On the other hand, link prediction is an equally important and widely studied task in graph learning. Many models like VGAE, LightGCN and GAT have emerged as the more powerful models for link prediction. Our work introduces a distinct weighted meta-learning framework with carefully designed attack modifications tailored specifically for adversarial link prediction tasks.

\section{Background}\label{sec:Background}
While the detailed approach is presented in a later section, this section provides the relevant background knowledge for this paper.

\subsection{Graph Adversarial Attack}
\begin{figure*}[h]
	\centering
	\includegraphics[scale=1]{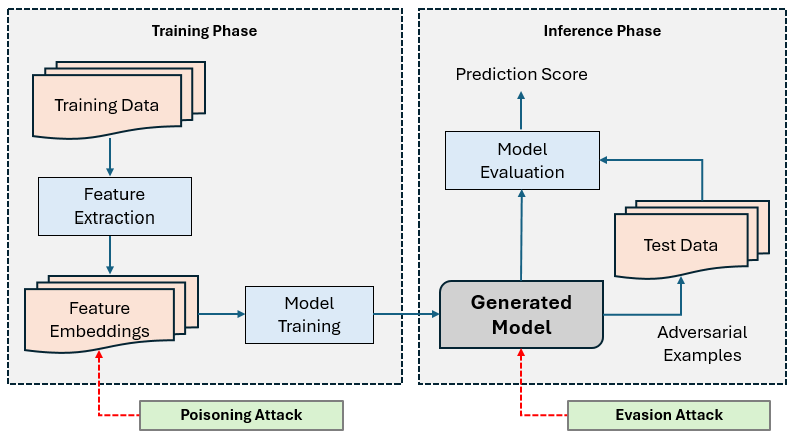}
	\caption{Adversarial attacks based on the target stage of the ML pipeline.}\label{attack_method}
\end{figure*}

In general, graph adversarial attacks deliberately modify the graph structure, node attributes, or related features to degrade the performance of a target model. Typical modifications include adding or deleting nodes or edges, altering link weights in weighted graphs, changing node features, or generating synthetic graphs. Such attacks can target machine-learning or deep-learning models at various stages of the learning pipeline. Based on the targeted stage, they can be broadly classified into two main categories (as illustrated in Fig.~\ref{attack_method}).

\begin{itemize}
    \item Evasion attack: This occurs during the model's testing (inference) phase. The attacker manipulates the input data during testing to induce the model to make incorrect predictions. Adversarial examples are typically generated to test the vulnerability of the target model.
    \item Poisoning attack: This occurs during the model's training phase. The attacker injects malicious data into the original training dataset, producing poisoned data that bias the model's learning process. The result is a dataset contaminated with malicious intent, used to train the target model. 
\end{itemize}

In the realm of graph adversarial attacks, the objectives pursued by attackers are pivotal in shaping their strategies. These adversarial attacks can be further classified depending on the attacker's goals and the context of the attack:
\begin{itemize}
    \item Untargeted attack: The objective is to reduce the overall prediction accuracy of the model in all instances.
    \item Targeted attack: The objective is to specifically reduce the prediction accuracy of the model for certain instance(s).
\end{itemize}

Furthermore, based on the consequences of the attack, graph adversarial attacks can be classified into two main categories:
\begin{itemize}
    \item Node classification attack: This type of attack aims to manipulate the predictions of the machine learning/deep learning model, leading it to misclassify node(s) into incorrect class(es). For example, it might force a node in one community to be classified into another community within a social network graph.    
    \item Link prediction attack: The objective of this attack is to manipulate the model's predictions regarding the likelihood of links between nodes. For example, the attack might cause the model to predict a relationship between two unrelated users in a social network graph, contrary to the actual scenario.
\end{itemize}

In this paper, our goal is to design a poisoning attack on link prediction tasks for untargeted scenarios.
\subsection{Graph Neural Networks}
\subsubsection{Variational Graph Auto-Encoders (VGAE)}
The variational graph auto-encoder proposed by Kipf et al.~\cite{kipf2016variational} is a attractive choice for link prediction tasks due to its capability to learn meaningful latent representations for graph data and to directly reconstruct the graph structure via the decoder. VGAE's objective is to maximize the Evidence Lower Bound (ELBO) $\mathcal{L}$ to enhance reconstruction accuracy while also regularizing the distribution of the latent space. Given the adjacency matrix $A$ and feature matrix $X$, the objective function of VGAE, derived from ELBO, can be expressed as:

\begin{equation}
\mathcal{L} = \mathbb{E}_{q(Z|X,A)}[\log p(A|Z)] - KL(q(Z|X,A) || p(Z))
\end{equation}
where $\mathbb{E}$ denotes the expected value and $KL$ represents the Kullback-Leibler divergence~\cite{kullback1951kullback} between the posterior $q$ and prior distribution $p$. KL divergence term acts as a regularizer, ensuring that the learned latent space is close to the prior distribution, which promotes better generalization and meaningful sampling.

\subsubsection{Graph Attention Network (GAT)}
The Graph Attention Network (GAT), introduced by Veličković et al.~\cite{velivckovic2017graph}, enhances graph representation learning by incorporating an attention mechanism. Unlike conventional graph convolutional networks that aggregate neighbor information uniformly, GAT allows nodes to assign different levels of importance to their neighbors. This is achieved by computing attention coefficients for each edge, which dynamically weigh the contribution of neighboring node features during the aggregation process. The core operation of a GAT layer can be expressed as:

\begin{equation}
\vec{h'}_{i} = \sigma\left(\sum{j \in \mathcal{N}_{i}} \alpha_{ij} \mathbf{W}\vec{h}_j\right)
\end{equation}
where $\vec{h'}_{i}$ is the updated feature vector for node $i$, $\sigma$ is a non-linear activation function, $\mathcal{N}_i$ represents the neighbors of node $i$, $W$ is a shared weight matrix for feature transformation, and $\alpha_{ij}$ is the normalized attention coefficient of node $j$ to node $i$. The attention coefficients are computed based on the features of the connected nodes, allowing the model to focus on the most relevant parts of the neighborhood. 

\subsubsection{LightGCN}
LightGCN, proposed by He et al.~\cite{he2020lightgcn}, is a simplified graph convolutional network designed specifically for recommendation and collaborative filtering. LightGCN simplifies the design by preserving neighborhood aggregation which is the essential part in GCN. It learns user and item embeddings by linearly propagating them on the user-item interaction graph. The aggregation at layer $k+1$ is defined as:

\begin{equation}
\mathbf{e}_u^{(k+1)} = \sum{i \in \mathcal{N}_u} \frac{1}{\sqrt{|\mathcal{N}_u|}\sqrt{|\mathcal{N}_i|}} \mathbf{e}_i^{(k)}
\end{equation}
where $\mathbf{e}_u^{(k)}$ and $\mathbf{e}_i^{(k)}$ are the embeddings of user u and item i at layer k, respectively, and the term with square roots is a symmetric normalization factor. The final embedding for a user is a weighted sum of its embeddings from all layers, which effectively combines information from its multi-hop neighbors. This simple linear propagation captures the collaborative filtering effect smoothly.



\subsection{Adversarial Attack with Meta-Learning}
Meta-learning, also referred to as \enquote{learn to learn}~\cite{finn2017model}, \cite{bengio2000gradient} is a widely used technology in the field of machine learning, recognized for its ability to facilitate efficient learning and rapid adaptation. Using knowledge acquired from various tasks, known as episodes, provides valuable insight into the model, enabling rapid adaptation to new or unseen tasks. Finn et al.~\cite{finn2017model} introduced a prominent meta-learning algorithm called Model-Agnostic Meta-Learning (MAML). MAML utilizes a fixed number of gradient descent steps on a meta-gradient, computed from various learning tasks (episodes), to achieve rapid parameter adaptation of the model for new tasks.

The concept of meta-learning can also be applied to graph adversarial attacks. As discussed in Section~\ref{sec:Related work}, attackers can leverage the meta-gradient obtained during training towards their target~\cite{zugner_adversarial_2019}. The meta-gradient is computed by unrolling the training process into multiple tasks, which can help the attack learn a more holistic picture of the model than looking into any certain phase in the training. By leveraging this meta-gradient, attackers gain a powerful road map, enabling them to manipulate graph data effectively and apply a powerful attack strategy. More specifically, the attacker can identify the node or links with the greatest impact during model learning and perform the precise attack instead of making random or naive perturbations on the graph. In this paper, we will take advantage of meta-learning to design our poisoning attack strategies.

\begin{figure*}[h]
	\centering
	\includegraphics[scale=0.4]{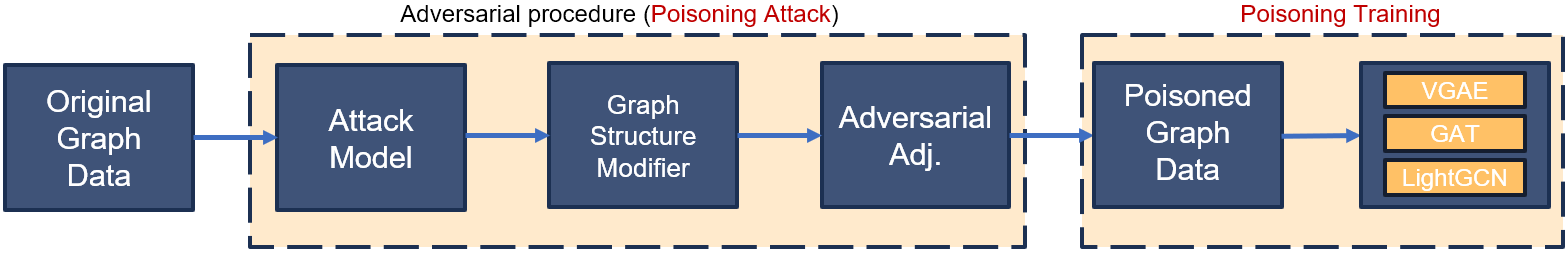}
	\caption{Framework of the proposed approach.}\label{framework}
\end{figure*}

\section{Methodology}\label{sec:Methodology}
This section outlines the methodology of the proposed approach, including its objectives, attack settings, and attack framework.

\subsection{Objective}
In the proposed adversarial attack approach, the attacker utilizes meta-learning with the weighted scheme during the training phase of the attack model to make subtle modifications to the unweighted graph structure (adjacency matrix, denoted as $Adj$). As depicted in Fig. \ref{framework}, these modifications generate poisoned data for the poisoning attack, which involves training the victim models with the poisoned data, named poisoning training.

The goal of this approach is to reduce the performance of the GNN models. In other words, our objective is to propose a model-agnostic poisoning attack to reduce the performance of link prediction models.

\subsection{Attack Setting and Threat Model}
\subsubsection{Attacker's Goal}
The attacker aims to increase the error rate of the link prediction models using the proposed poisoning attack. By introducing malicious and inconspicuous changes to the graph data, the attacker tries to make the learned link prediction models inherently biased after training on the poisoning data. For example, the goal is to induce the model to predict a link between two unrelated nodes, which should not be predicted.

\subsubsection{Attacker's Knowledge}
In the original machine learning and deep learning process, there are typically three components: data, model, and training pipeline. In our approach, we assume that the attacker only has limited knowledge.
\begin{itemize}
    \item Graph Data Knowledge: The attacker has full access to the graph structure (adjacency matrix), node features, and node labels.
    \item Model Knowledge: The attacker does not know the exact model architecture, such as the number of layers and hyperparameters of the model. 
    \item Training Pipeline Knowledge: The attack has a general understanding of when the model is trained or updated. However, they do not have control or alter the original training procedure (i.e., the optimizer, learning rate). 
\end{itemize}

\subsubsection{Attacker's Capacity}
In the proposed poisoning attack, the attacker has the ability to manipulate graph data with a limited budget:
\begin{itemize}
    \item Graph Structure Manipulation: The attacker can add or remove links in the adjacency matrix. The other modification to the graph data is not allowed.
    \item Attack Budget: The attacker has a budget that restricts how many links can be added or removed. This budget typically scales with the size of the graph to allow sufficient modifications to be made on large graphs. By limiting the attack budget, the attacker can achieve inconspicuous modifications. 
\end{itemize}

\subsubsection{Threat Model}
Leveraging the attacker's knowledge and capacity, the attacker selectively adds or removes links within the allowed budget to generate poisoning data via their own framework (Section~\ref{sec:attackframe}). These poisoned data distort the structural patterns of the graph, causing the link prediction models to learn misleading latent representations. In other words, the attacker relies on subtle but carefully chosen structural modification on the graph data to achieve their goal of inducing incorrect link prediction.

\subsection{Attack Framework} \label{sec:attackframe}
Fig. \ref{framework} provides an overview of our approach. It begins with the original graph data (original $Adj$ and original features) as input and generates poisoned graph data (poisoned $Adj$ and original features) for poisoning training. Our adversarial procedure consists of two primary components that operate sequentially to produce the poisoned graph data to produce poisoned data.

\begin{itemize}
    \item Attack model: The attacker employs a surrogate model as an attack model to train on the original graph data. During the surrogate model training, the attacker utilizes the weighted meta-learning to construct the meta-gradient. This meta-gradient helps the attacker to identify effective modifications to the graph structure.
    \item Graph structure modifier: Following the training of the attack model, the attacker employs a graph structure modifier to adjust the original adjacency matrix (original $Adj$) and create an adversarial adjacency matrix (adversarial $Adj$) based on meta-gradient. Subsequently, the attacker combines the adversarial $Adj.$ with the original graph features to generate poisoned graph data for poisoning training.
\end{itemize}

\begin{algorithm}[h]
\caption{Adversarial Attack on VGAE using \textit{Attack Model} with Meta-learning}\label{attackModel}
\LinesNumbered
\KwIn{Attack budget $\Delta$, graph data $G(A,X)$, number of epoches $k$, attack model $M$, weight mechanism $W$}
\KwOut{Adversarial adjacency matrix $A_{\text{adv}}$}
Initialize model parameters $\theta_{0}$ for $M$\;

\For{$i = 0$ to $k$}{
    Train $M$ and compute training loss $L$\;
    Compute the gradient $ \nabla $ with respect to $\theta_{i}$ based on $L$\;
    Compute attack loss $\widetilde{L}_{i}$\;
    Select $w_{i}$ from $W$\;
    $\widetilde{L} = \widetilde{L} + w_{i} \times \widetilde{L}_{i}$\;
    Back propagation on $\nabla$: update  $\theta_{i} \rightarrow \theta_{i+1}$\;
}

Compute the meta-gradient $\widetilde{\nabla}$ with respect to attack target based on attack loss ($\widetilde{L}$)\;

$A_{adv} =$ Adversarial modification ($\widetilde{\nabla}, A, \Delta$)\;

\Return $A_{adv}$\;
\end{algorithm}

    
     

\begin{algorithm}[h]
\caption{Adversarial Modification (Graph Structure Modifier)}\label{graphStructureModifier}
\LinesNumbered
\KwIn{Gradient matrix $ \widetilde{\nabla} $, adjacency matrix $A$, attack budget $\Delta$}
\KwOut{Adversarial adjacency matrix: $A_{\text{adv}}$}
Initial $A_{adv}= A$\\
Symmetry the gradient matrix $ \widetilde{\nabla} $ = ($ \widetilde{\nabla} $ + $ \widetilde{\nabla}^{T}$)/2 \\
\While{$\Delta > 0$}{
    Get the index of the largest magnitude in $ \widetilde{\nabla} $:$(i, j)$ = argmax($| \widetilde{\nabla} |$);\\
    \If{$ \widetilde{\nabla}_{ij} > 0$ and $A_{ij}=0$}{
     $A_{adv}(i,j)=1$\;
     $\Delta$ =  $\Delta$ - 1\; 
    } 
    \If{$ \widetilde{\nabla}_{ij} < 0$ and $A_{ij}=1$}{
    $A_{adv}(i,j)=0$\;
    $\Delta$ =  $\Delta$ - 1\; 
    } 
    Set position (i, j) immutable\;
}
\Return $A_{adv}$\;
\end{algorithm}

\subsection{Attack Strategy}
Algorithms \ref{attackModel} and \ref{graphStructureModifier} outline the step-by-step process to generate the adversarial adjacency matrix through the attack model and make adversarial modifications using the graph structure modifier.

\subsubsection{Attack Model}
The primary purpose of the attack model is to extract information by mimicking the original process since the attacker lacks access to the original model and the training process. The essence of the attack model (Algorithm \ref{attackModel}) lies in leveraging weighted meta-learning during its training phase, enabling the attacker to gain a comprehensive understanding of the process. The attack model is trained on the original graph data using weighted cross-entropy (Algorithm 1, lines 3 and 4), defined as:

\begin{equation}
    L = \sum_{ij} -w{A_{ij}} \ln(\hat{A}_{ij}) - (1 - A_{ij}) \ln(1 - \hat{A}_{ij})
\end{equation}
where $w$ is the weight for weight cross-entropy, $w = (N^2 - \sum_{ij} A_{ij}) / \sum_{ij} $, $N$ represent the number of nodes, $A_{ij}$ is the label in original $Adj$, and $\hat{A}_{ij}$ is the predicted label. 

The attacker minimizes this training loss with respect to the attack model parameter to optimize the performance of link prediction task. In addition, the attacker computes the attack loss in each training epoch (Algorithm 1, line 5), defined as:

\begin{equation}
    \tilde{L} = -w{Y}_t \ln(\tilde{A}_t) - (1 - {Y}_t) \ln(1 - \tilde{A}_t)
\end{equation}
where $w$ is the weight for weight cross-entropy, $w = (N^2 - \sum_{ij} A_{ij}) / \sum_{ij}$, $N$ represent the number of nodes, ${Y}_t$ is the label in original $Adj$, and $\tilde{A}_t$ is the predicted label.

The attacker leveraging the meta-learning idea to aggregate the attack loss during each epoch (Algorithm 1, line 6). The key part of the attack model is that the attacker considers each training epoch as a task (episode) and intends to learn from all tasks (epoches). Instead of looking into a specific phase in the training process, meta-learning helps the attacker to capture the overall information of the whole training. The attacker tries to maximize the attack loss with respect to the graph structure ($Adj$) to conduct the effective attack on the graph structure. At the end of the training, the attacker computes the meta-gradient from the attack loss for further processing (Algorithm 1, line 8). 

\subsubsection{Weighted Scheme}
Meta-gradient is constructed to guide the subsequent adversarial modifications. Hence, it is critical to form an informative and effective meta-gradient. To enhance its effectiveness, we propose three distinct weighting schemes that assign varying importance to gradient information from each training epoch (Algorithm 1, line 7). These schemes determine how each epoch’s gradient contributes to the final aggregated meta-gradient:
\begin{itemize}
    \item Magnitude-based Weighting:\\
     Magnitude-based weighting builds on fundamental optimization principles and relates to influence functions ~\cite{koh2017understanding}. The gradient’s magnitude directly measures the loss function’s sensitivity to input changes. This scheme assumes that larger gradients indicate epochs of greater sensitivity to structural perturbations and thus higher importance. Accordingly, the gradient in epoch $i$ is weighted by its L2 norm to capture this effect:
    \begin{equation}
        w_{i} = \left\| \widetilde{\nabla} \tilde{L}_{i} \right\|_2
    \end{equation}
 
    \item Performance-based Weighting:\\
    The core idea of boosting algorithms~\cite{freund1997decision} like AdaBoost indicates that the final prediction is a weighted combination of models, with more accurate models receiving a greater say. Similarly, our performance-based scheme uses the attack model's own performance as a signal for attack reliability. This scheme links the epoch importance directly to the empirical performance of the attack model. The weight for each epoch is related to its Average Precision (AP) score on the validation data of link prediction task. It assumes that gradient is more informative when the model has a good performance (understanding) of the validation data. The weight for epoch $i$ is calculated as:\
    \begin{equation}
        w_{i} = AP_{i}
    \end{equation}

    \item Linear-based Weighting:\\
    Inspired by curriculum learning~\cite{bengio2009curriculum}, where models learn more effectively when data is introduced from simple to complex, our linear weighting adopts a similar philosophy. We hypothesize that gradients from later epochs carry greater attack-relevant information. Thus, weights increase linearly with training progress. As the model captures the basic data structure in early epochs, this scheme prioritizes gradients from later stages. The weight for epoch $i$ is normalized by the total number of epochs $k$ and computed as:\
    \begin{equation}
        w_{i} = i/k
    \end{equation}
    
\end{itemize}

\subsubsection{Adversarial modification}
The meta-gradient obtained from the attack model training provides the attack direction for the graph structure modifier to apply the adversarial modification on the graph. Specifically, the values in the meta-gradient matrix represent their impact during model training, with larger magnitude (absolute values) indicating greater effectiveness. The largest magnitude of the meta-gradient points out the most effective link positions (Algorithm \ref{graphStructureModifier}, line 4). Based on this, the attacker continues to modify the corresponding positions of the original $Adj.$ to acquire the adversarial $Adj.$ (Algorithm \ref{graphStructureModifier}, lines 4-11) until the attack budget is exhausted. Since the graph structure in an unweighted graph is discrete, the graph structure modifier is limited to two actions: adding or removing links in the original $Adj.$ matrix. This involves altering the values of the adjacency matrix to either 0 (to remove a link) or 1 (to add a link). Furthermore, the attack budget imposes constraints on the total number of modifications allowed. By carefully regulating the quantity and nature of these modifications, the attack can execute unnoticeable poisoning attacks on the graph data that are hard to detect.

\section{Experimental Results}\label{sec: Experimental Evaluation}
In this section, we outline our experimental design and analyze the results obtained from our approach.

\subsection{Experimental Setup}
To evaluate the effectiveness of our proposed attack method, we implemented the entire architecture using PyTorch. Our experiments are designed to answer the following questions:

\begin{itemize}
    \item Does our approach effectively reduce link prediction performance?
    \item What is the most effective weighted scheme?
    \item How does our approach compare to the state-of-the-art baseline methods?
    \item What is the impact of the attack budget on our approach?
\end{itemize}

To answer these questions, we deployed VGAE as the attack surrogate model and evaluated the effectiveness of our poisoning attack approach in various graph-based use cases, including citation network (Cora), social network (Twitch) and recommendation network (MovieLens). Each use case is evaluated on a state-of-the-art link prediction model with the corresponding dataset. By training link prediction models separately on the original and poisoned graph data, we can observe any disparities in link prediction performance, indicating the effectiveness of our attack. In addition, we compared the effectiveness of our approach with other state-of-the-art methods, namely PGD~\cite{xu2019topology}, DICE~\cite{waniek2018hiding}, IGA~\cite{chen2018link}, CLGA~\cite{zhang2022unsupervised} and MetaBase~\cite{zugner_adversarial_2019}. The implementations for PGD, DICE, and MetaBase were adapted from the Graph Reliability Toolbox~\cite{GreatX}, and CLGA was sourced from the authors' public repository. As an official implementation for IGA was unavailable, we implemented it ourselves based on the description in the original paper.

All experiments were conducted on a server equipped with an NVIDIA 3090 GPU. We utilize a 2 layer VGAE (with 32 and 16 hidden units) as the attack surrogate model to generate adversarial modifications to the graph structure. The effectiveness and transferability of these attacks are then evaluated against three link prediction victim models: 1) a two-layer Graph Attention Network (GAT) with 32 and 16 hidden units, 2) a two-layer VGAE with 32 and 16 hidden units, and 3) a three-layer LightGCN with a 32-dimensional embedding space. As described in the previous section, the adversarial modification is constrained by the attack budget, which we treat as a key experimental parameter. To keep the attacks inconspicuous, we limit the budget to a small range, between 1\% and 5\% of the total links in the dataset, when generating poisoned graphs. For each experiment, we train both the original and the poisoned models using their respective datasets and evaluate them on the same test set to assess link prediction performance.

\subsection{Experimental Dataset and Models}
Since the proposed approach requires to access the full adjacency matrix to gain a global perspective of the graph structure to guide the attack most effectively, the primary bottleneck is its space complexity ($O(N^{2})$). This constraint creates a trade-off between our attack's global effectiveness and its scalability. To prioritize a rigorous validation of the novel attack algorithm itself, we selected three widely used graph datasets in the literature to evaluate the performance of our proposed attack approach. These datasets represent the use cases in citation network, social network, and recommendation network. 

\begin{itemize}
    \item Cora (citation network): The Cora dataset~\cite{mccallum2000automating} contains scientific publications that are categorized into 7 classes. It has 2,708 nodes and 5,429 links. Each node is described by a word vector that indicates the presence of certain words in the publication from a predefined dictionary. This dataset is used to evaluate the GAT model.
    \item Twitch (social network): The Twitch Gamers dataset~\cite{rozemberczki2019multiscale} is a social network of users from the Twitch streaming platform. The version for English-speaking users contains 7,126 nodes and 35,324 links, where nodes are users and links represent mutual friendships. Node features are created based on user profiles and streaming activities. This dataset is used to evaluate the VGAE model.
    \item MovieLens (recommendation network): The MovieLens dataset~\cite{harper2015movielens} consisting of 600 users and 9,000 movies. The graph is represented as a bipartite graph, where edges connect users to the movies they have rated. The features are represented by learnable embeddings to evaluate the LightGCN model.
\end{itemize}

\begin{figure*}[h]
\centering
	\begin{subfigure}{.32\textwidth}
		\includegraphics[width=\textwidth]{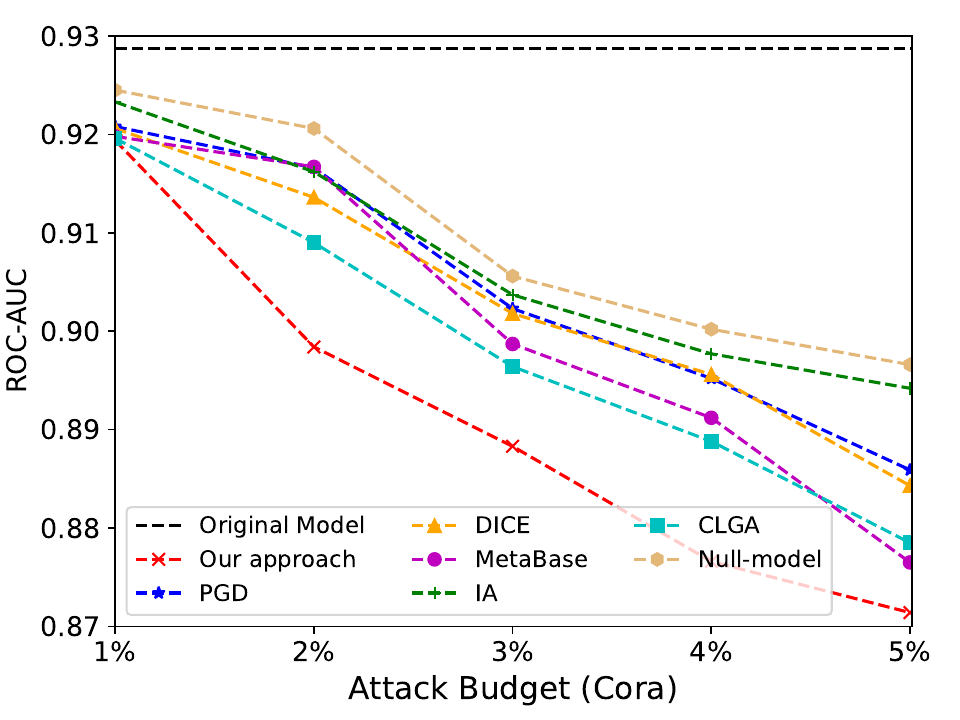}
		\caption{Cora dataset}
        \label{ROC_cora}
	\end{subfigure}
	\begin{subfigure}{.32\textwidth}
		\includegraphics[width=\textwidth]{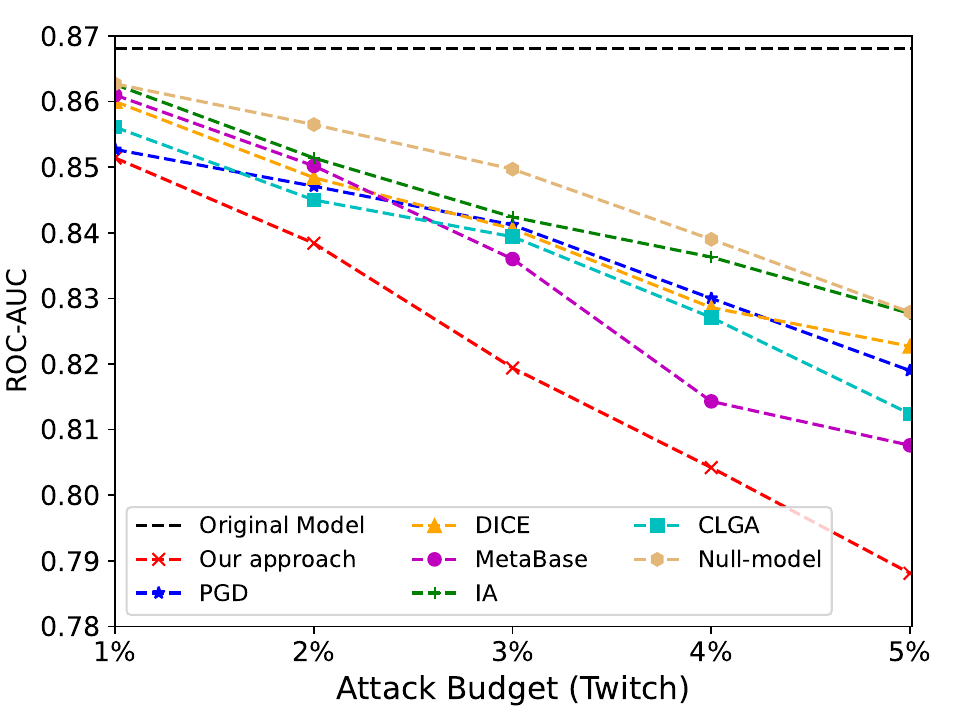}
		\caption{Twitch dataset}
        \label{ROC_twitch}
	\end{subfigure}
	\begin{subfigure}{.32\textwidth}
		\includegraphics[width=\textwidth]{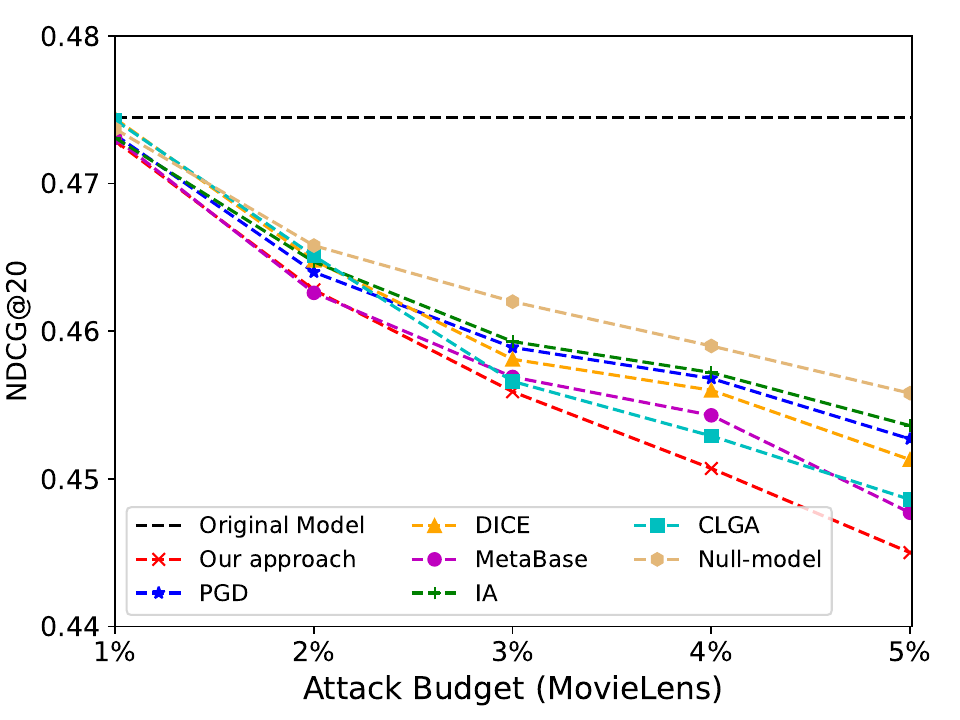}
		\caption{MovieLens dataset}
        \label{NDCG_movielens}
	\end{subfigure}
	\caption{Comparsion using ROC-AUC and NDCG@20 metrics}
    \label{Fig: ROC}
\end{figure*}
\begin{figure*}[h]
\centering
	\begin{subfigure}{.32\textwidth}
		\includegraphics[width=\textwidth]{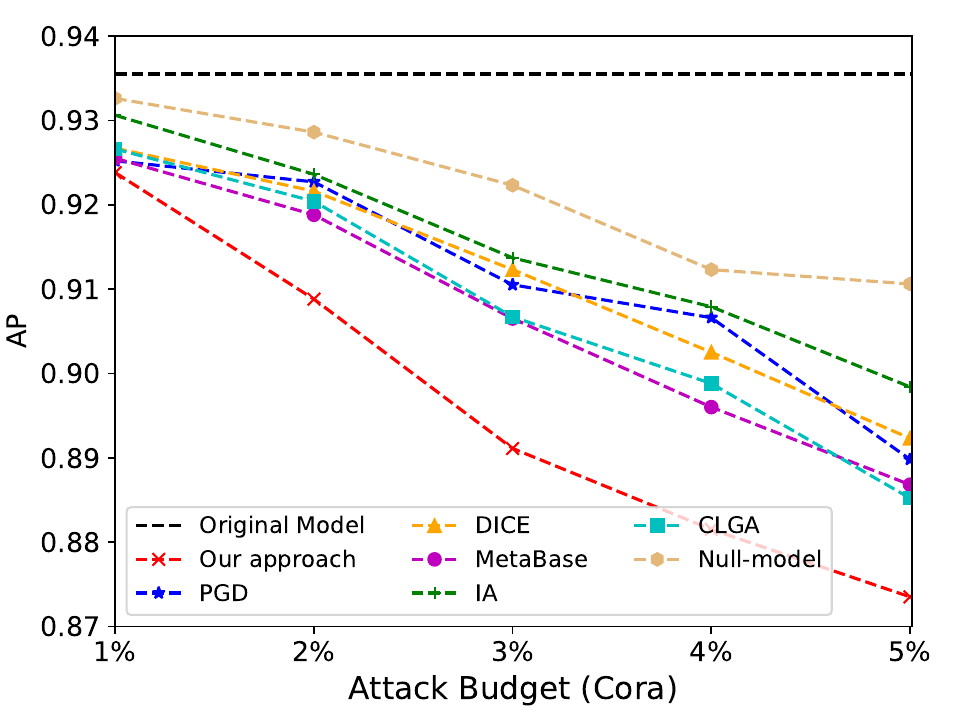}
		\caption{Cora dataset}
        \label{AP_cora}
	\end{subfigure}
	\begin{subfigure}{.32\textwidth}
		\includegraphics[width=\textwidth]{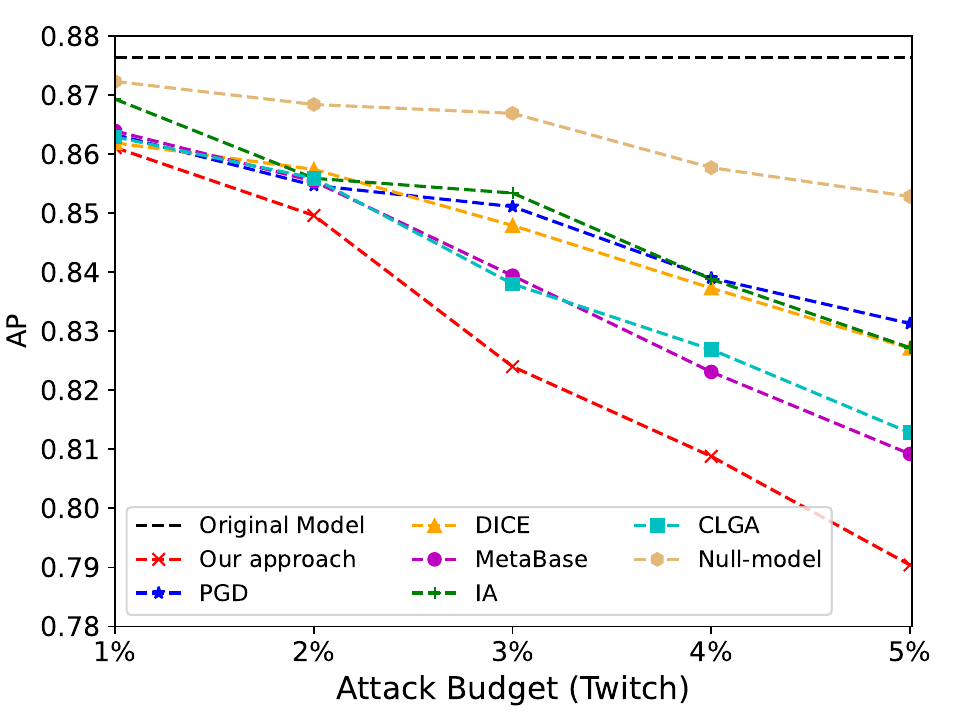}
		\caption{Twitch dataset}
        \label{AP_twitch}
	\end{subfigure}
	\begin{subfigure}{.32\textwidth}
		\includegraphics[width=\textwidth]{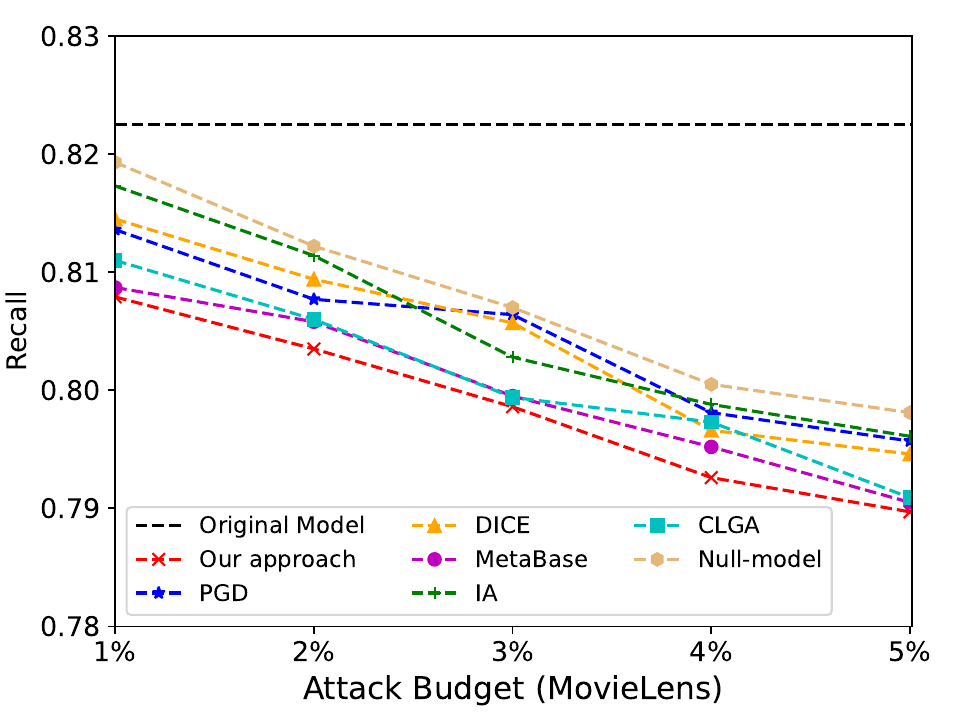}
		\caption{MovieLens dataset}
        \label{RECALL_movielens}
	\end{subfigure}
	\caption{Comparsion using AP and Recall metrics}
    \label{Fig: AP}
\end{figure*}

\subsection{Evaluation Metric}
\begin{itemize}
    \item ROC-AUC and AP Score: For citation and social networks, we use the ROC-AUC (Area Under the ROC Curve) and the Average Precision (AP) score to evaluate the performance of the model after the poisoning attack. ROC-AUC score provides a comprehensive understanding of the model's ability to distinguish between positive and negative links. It evaluates the performance of the model across all possible thresholds and is more resistant to class imbalance compared to accuracy, making it a robust measure. As graph data are usually considered imbalance data, the AP score emphasizes the precision-recall trade-off of the model. By comparing clean model training with poisoning training performance, the effectiveness of the poisoning attack can be clearly demonstrated, highlighting its impact on the robustness and performance of the model.
    
    \item NDCG@k and recall: For the recommendation network, the evaluation focuses on the quality of the top-K item ranking rather than the simple prediction of links. We employ Normalized Discounted Cumulative Gain (NDCG@k) and Recall for evaluation. NDCG@k evaluates the quality of the ranking itself. It is a position-aware metric that assigns greater importance to relevant items placed higher on the list, making it highly sensitive to recommendation order. Recall measures the fraction of relevant items from the test set that are successfully included in the top K recommended list. A significant drop in these top K ranking metrics after our poisoning attack demonstrates the attack's effectiveness in degrading the model's ability.
\end{itemize}

\begin{table}[h!]
\centering
\caption{Performance drop on the Cora dataset}
\label{tab:cora_combined}
\resizebox{\columnwidth}{!}{%
\begin{tabular}{@{}llcccc@{}}
\toprule
\textbf{Metric} & \textbf{Attack Budget} & \textbf{Uniform} & \textbf{Performance} & \textbf{Magnitude} & \textbf{Linear} \\
\midrule
\multirow{3}{*}{$\Delta$ROC} & 1\%   & 0.0076 & 0.0101 & 0.0116 & 0.0093 \\
                          & 2.5\% & 0.0388 & 0.0401 & 0.0326 & 0.0403 \\
                          & 5\%   & 0.0526 & 0.0570 & 0.0487 & 0.0573 \\
\midrule
\multirow{3}{*}{$\Delta$AP} & 1\%    & 0.0108 & 0.0125 & 0.0142 & 0.0117 \\
                         & 2.5\%  & 0.0426 & 0.0441 & 0.0359 & 0.0444 \\
                         & 5\%    & 0.0562 & 0.0620 & 0.0528 & 0.0620 \\
\bottomrule
\end{tabular}
}
\end{table}

\begin{table}[h!]
\centering
\caption{Performance drop on the Twitch dataset}
\label{tab:twitch_combined}
\resizebox{\columnwidth}{!}{%
\begin{tabular}{@{}llcccc@{}}
\toprule
\textbf{Metric} & \textbf{Attack Budget} & \textbf{Uniform} & \textbf{Performance} & \textbf{Magnitude} & \textbf{Linear} \\
\midrule
\multirow{3}{*}{$\Delta$ROC} & 1\%   & 0.0020 & 0.0056 & 0.0058 & 0.0067 \\
                          & 2.5\% & 0.0353 & 0.0416 & 0.0326 & 0.0487 \\
                          & 5\%   & 0.0504 & 0.0552 & 0.0456 & 0.0804 \\
\midrule
\multirow{3}{*}{$\Delta$AP} & 1\%    & 0.0038 & 0.0136 & 0.0068 & 0.0152 \\
                         & 2.5\%  & 0.0417 & 0.0426 & 0.0405 & 0.0523 \\
                         & 5\%    & 0.0508 & 0.0567 & 0.0477 & 0.0859 \\
\bottomrule
\end{tabular}
}
\end{table}

\begin{table}[h!]
\centering
\caption{Performance drop on the MovieLens dataset}
\label{tab:movielens_combined}
\resizebox{\columnwidth}{!}{%
\begin{tabular}{@{}llcccc@{}}
\toprule
\textbf{Metric} & \textbf{Attack Budget} & \textbf{Uniform} & \textbf{Performance} & \textbf{Magnitude} & \textbf{Linear} \\
\midrule
\multirow{3}{*}{$\Delta$NDCG@20} & 1\%   & 0.0010 & 0.0015 & 0.0008 & 0.0016 \\
                              & 2.5\% & 0.0156 & 0.0169 & 0.0073 & 0.0186 \\
                              & 5\%   & 0.0210 & 0.0171 & 0.0154 & 0.0295 \\
\midrule
\multirow{3}{*}{$\Delta$Recall} & 1\%    & 0.0108 & 0.0123 & 0.0006 & 0.0146 \\
                             & 2.5\%  & 0.0198 & 0.0187 & 0.0092 & 0.0239 \\
                             & 5\%    & 0.0219 & 0.0231 & 0.0123 & 0.0328 \\
\bottomrule
\end{tabular}%
}
\end{table}

\begin{figure*}[h!]
\centering
	\begin{subfigure}{.49\textwidth}
		\includegraphics[width=\textwidth]{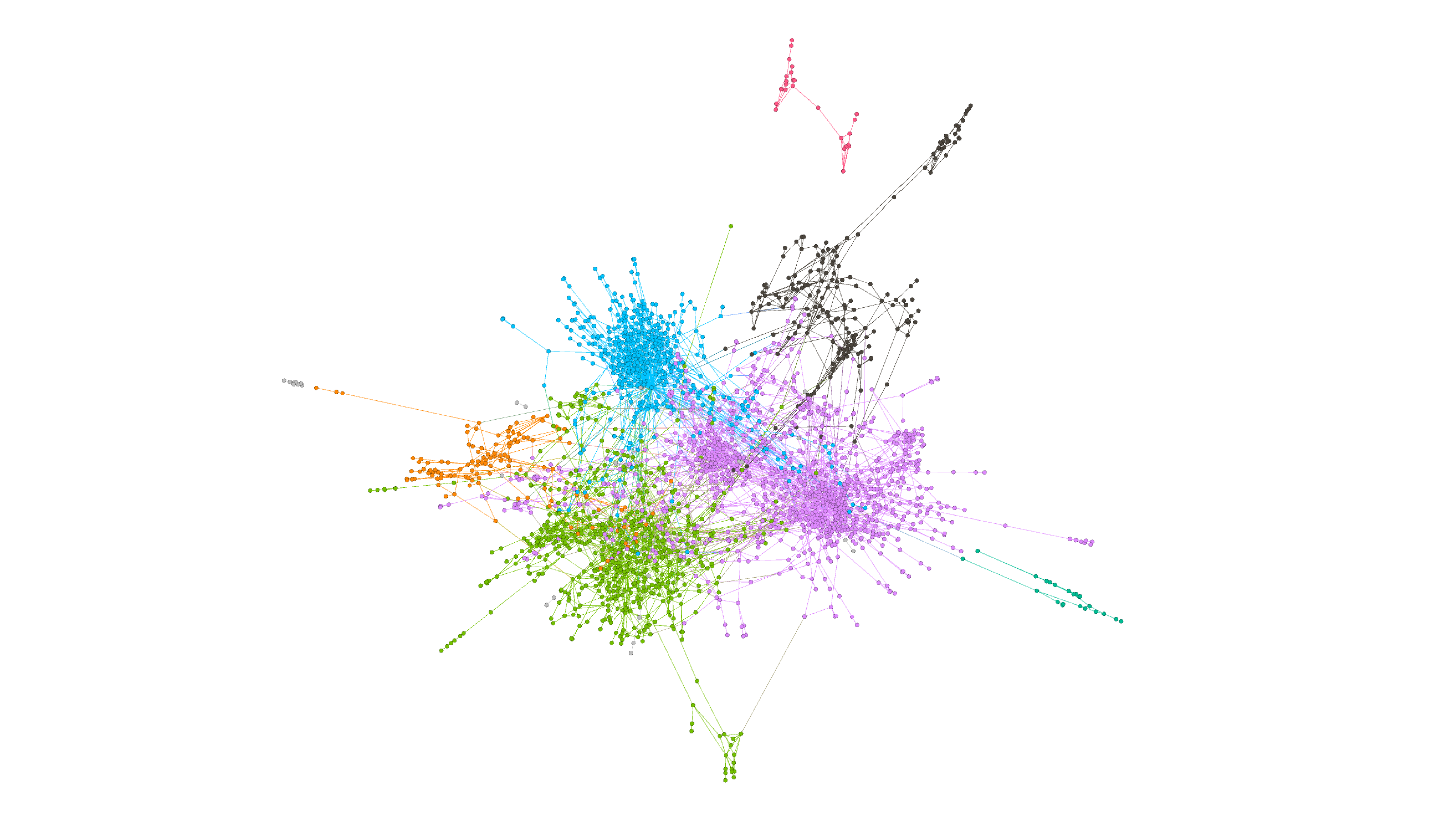}
		\caption{Original Graph}
        \label{visualization of original graph}
	\end{subfigure}
	\begin{subfigure}{.49\textwidth}
		\includegraphics[width=\textwidth]{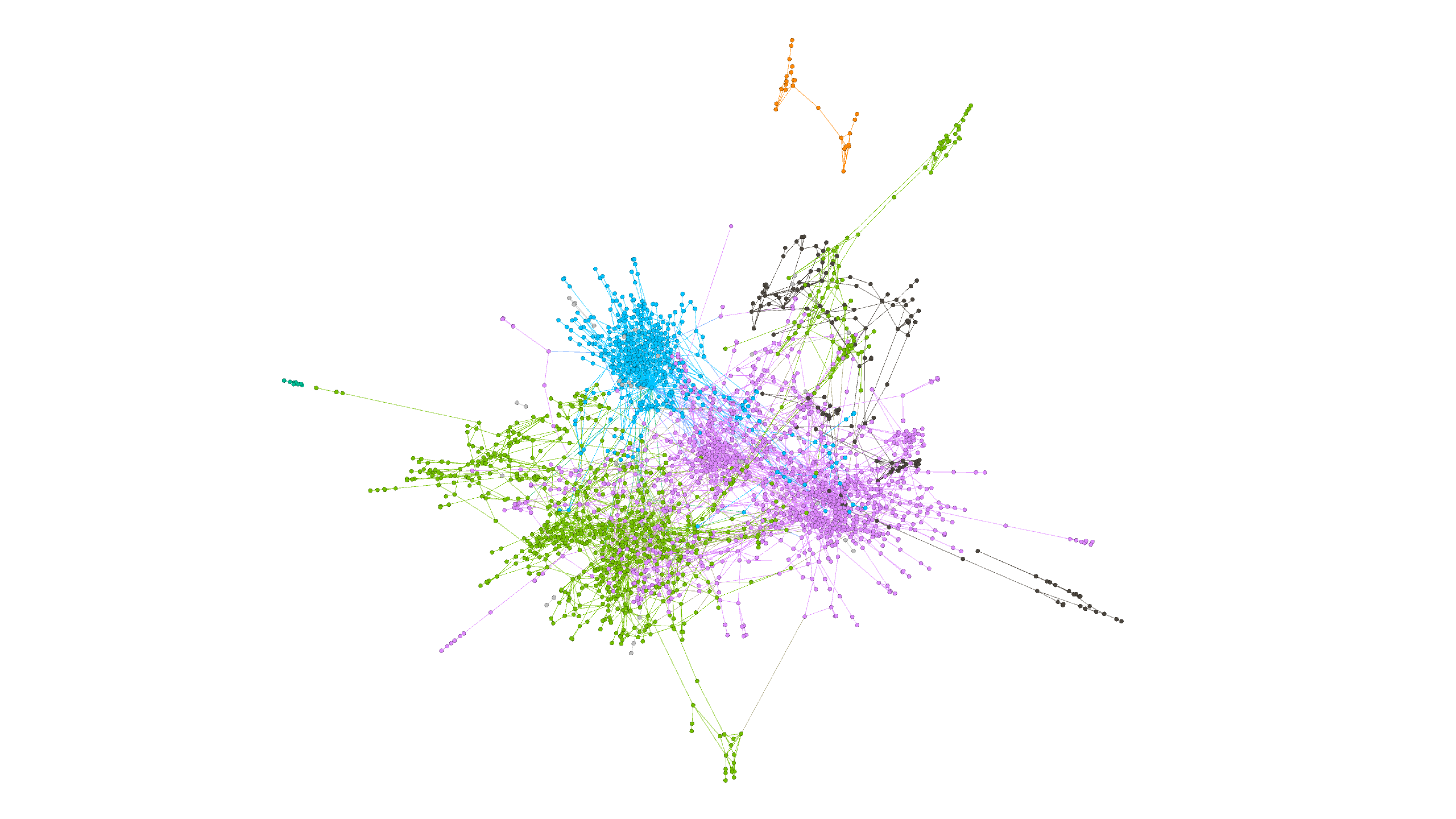}
		\caption{Poisoned Graph (5\%)}
        \label{visualization of poisoned graph}
	\end{subfigure}
	\begin{subfigure}{.49\textwidth}
		\includegraphics[width=\textwidth]{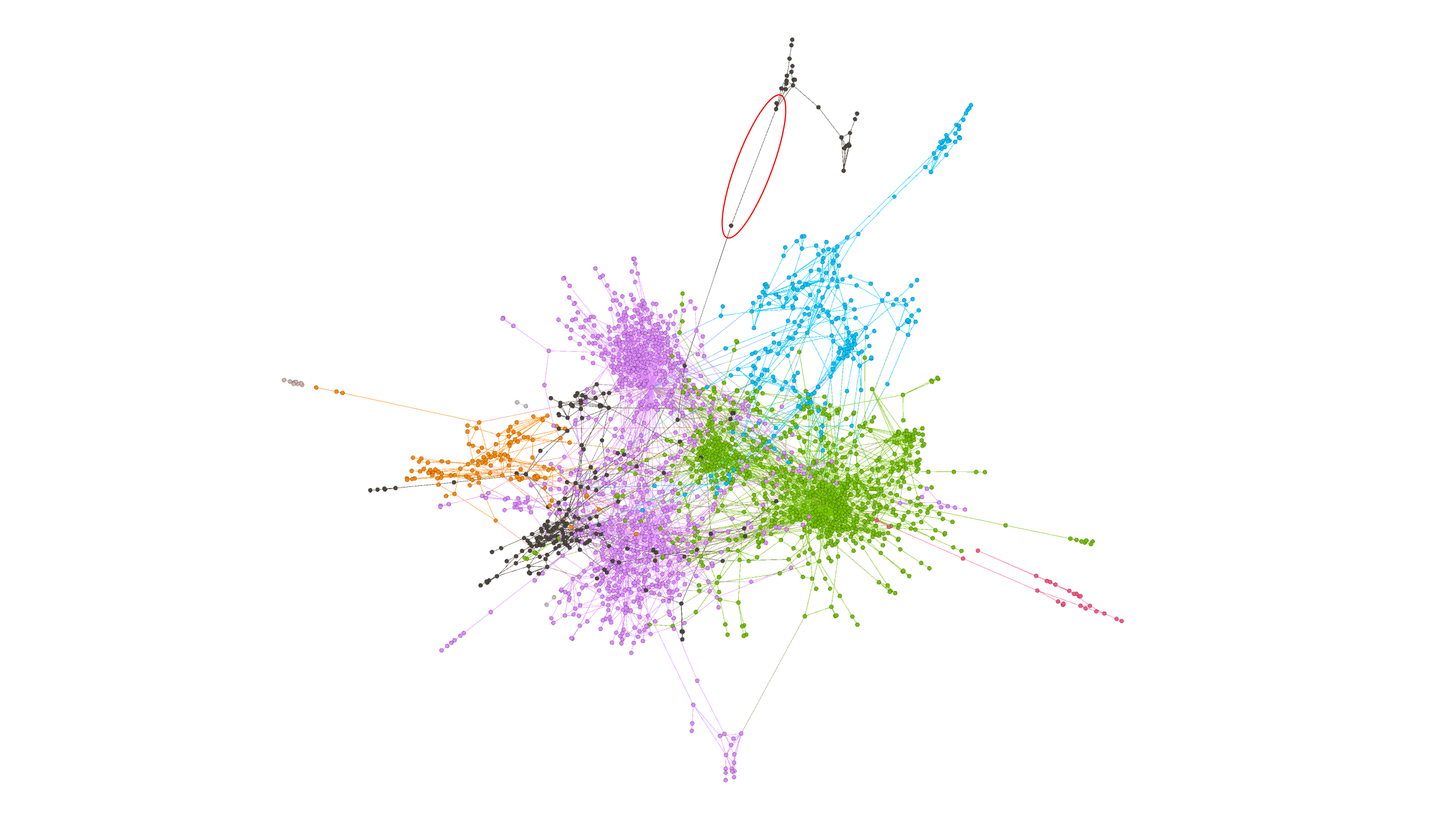}
		\caption{Poisoned Graph (10\%)}
        \label{visualization of poisoned graph}
	\end{subfigure}
	\begin{subfigure}{.49\textwidth}
		\includegraphics[width=\textwidth]{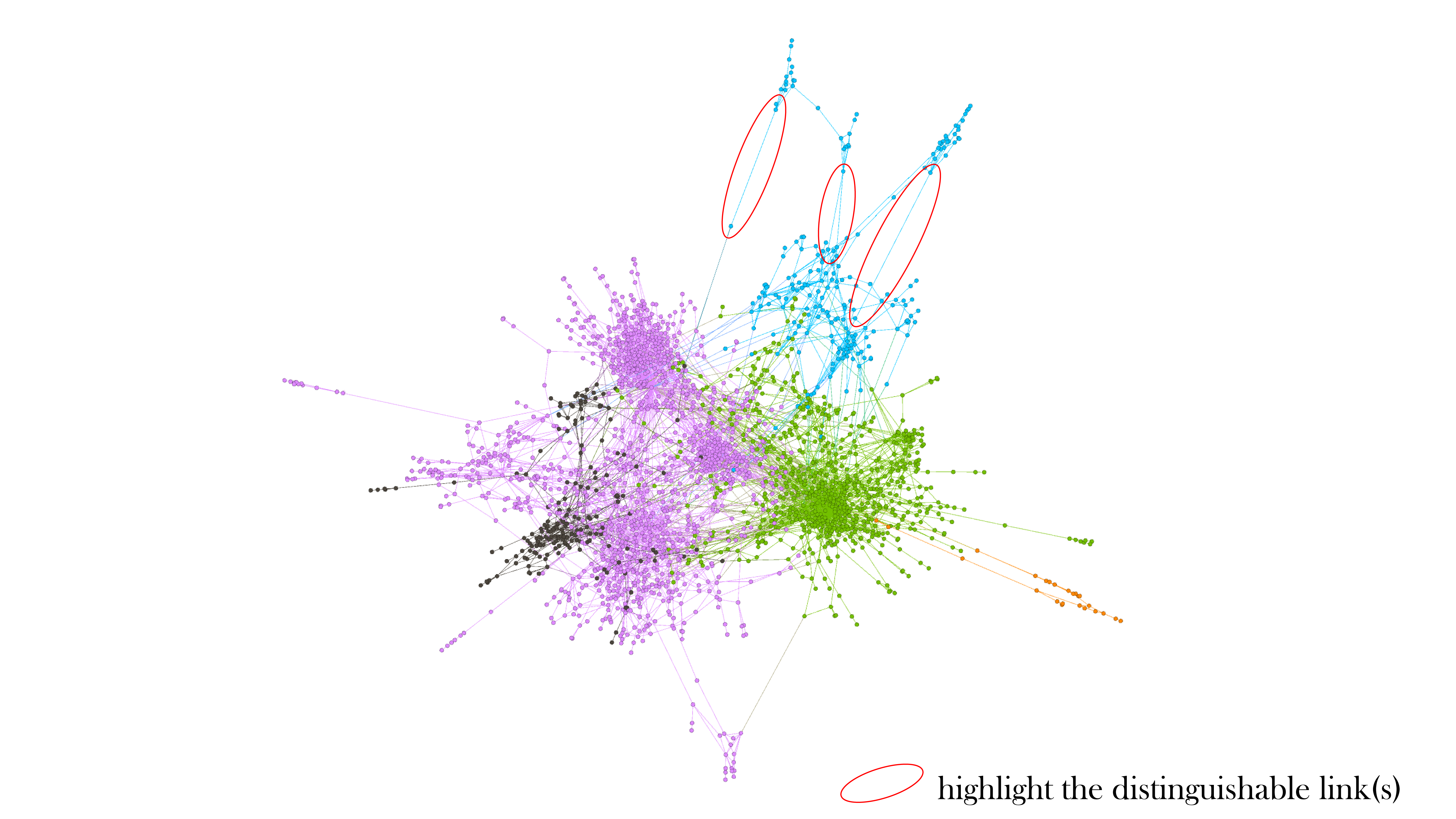}
		\caption{Poisoned Graph (20\%)}
        \label{visualization of poisoned graph}
	\end{subfigure}
 \caption{Visualization Examples of original graph and poisoned graph resulting from the proposed attack.}
 \label{visulization}
\end{figure*}

\begin{table*}[h!]
\centering   
\begin{tabular}{|c|c|c|c|c|}
    \hline
    Graph Statistic & Original Graph &  Poisoned Graph (5\%) & Poisoned Graph (10\%)&  Poisoned Graph (20\%) \\ \hline
    Number of Nodes & 2708  & 2708 & 2708 & 2708 \\ \hline
    Number of Edges & 4488  & 4656  & 4986  & 5358\\ \hline
    Avg. Degree & 3.42 & 3.56 & 3.80 & 4.15 \\ \hline
    Graph Density & 0.001 & 0.001 & 0.001 & 0.002\\ \hline
    Network Diameter & 21  & 21 & 18 & 14 \\ \hline
    Avg. Clustering Coefficient &0.24 & 0.25  & 0.27 & 0.30 \\ \hline
    Avg. Path Length & 6.79 & 6.42 & 5.80 & 5.00 \\ \hline
    \end{tabular}
    \caption{Comparison of statistics of the two graphs.}
    \label{statistic}
\end{table*}

\subsection{Effectiveness Analysis}
\subsubsection{Evaluation of weighted scheme}
To evaluate the effectiveness of our proposed weighted scheme, we conducted extensive experiments on benchmark datasets under proposed weighted schemes to observe the performance drop from clean training to poisoning training. The results are summarized in Tables \ref{tab:cora_combined}, \ref{tab:twitch_combined}, and \ref{tab:movielens_combined}. These experiments are designed to compare our three proposed weighting schemes against a standard baseline and to analyze their relative impact. We establish a baseline for comparison using a uniform weighting scheme which is the general aggregation approach for adversarial meta-learning. As results shown in three tables, our proposed weighted schemes (Performance, Magnitude, and Linear) significantly outperform this baseline. For instance, on the Twitch dataset with a 5\% attack budget, the linear-based scheme achieves a $\Delta$AP of 0.0859, a substantially more effective result than the uniform scheme's 0.0508. This clearly demonstrates the advantage of using adversarial meta-learning with proposed weighted schemes.

Among the three proposed schemes, the linear-based scheme consistently emerges as the most impactful and powerful weighted scheme. It achieves the highest performance drop in the vast majority of experimental settings, showcasing its robustness across different data domains and evaluation metrics. The performance-based scheme also proves to be highly effective, delivering results that are often comparable to the linear scheme. In contrast, the magnitude-based scheme, while still superior to the uniform baseline, generally yields the smallest performance drop of our three proposed methods. The findings confirm that attack guided by the proposed weighting schemes can compromise far more effectively than an unguided, uniform attack, especially those using linear-based and performance-based weighting schemes.

\subsubsection{Evaluation of poisoning attack}
To evaluate the effectiveness of the proposed poisoning attack, we benchmarked it using the most impactful weighted scheme (linear-based) against several state-of-the-art adversarial attack baselines, shown in Fig. \ref{Fig: ROC} and Fig. \ref{Fig: AP}.  In the experiments, we used two baselines, the original model and the null model~\cite{shang2014construction}. The original model represents the performance of the link prediction on the original graph data. It is unaffected by the attack budget as it is clean data. The effectiveness of the attack methods is evaluated by observing the drop on the evaluation metrics. The null model serves as a reference point to evaluate whether the approach is capable of highlighting the vulnerabilities of the graph data. All approaches demonstrate their effectiveness compared to the original model and null model. As shown in the figures, the proposed approach consistently achieves a significant drop in the victim model's performance, slightly surpassing other state-of-the-art approaches. As the attack budget increases, the performance of the models experiences a significant drop. In particular, our method achieves a lower performance score while requiring a smaller attack budget. In conclusion, the comparison provides evidence for the superiority of our proposed method, demonstrating its advanced capability to identify and exploit critical vulnerabilities in link prediction models.


\subsection{Graph Visualization}
Fig. \ref{visulization} shows a visualization example of the original graph and the poisoned graph under different attack budgets (ratio of the total number of edges) resulting from the proposed attack using the Cora dataset. All visualizations were generated using Gephi with the same layout setting and fixed node coordinates. To better present the local structure of the graph data, we also applied the community detection results on the visualization example. The nodes are colored according to their community labels, with the labels ranked by community size from largest to smallest as follows: purple, green, blue, black, orange, and red. The purpose of visualizing these graphs is to investigate how the proposed attack affects the original graph. In other words, we do not expect the proposed attack to produce conspicuous modifications to the original graph. As we mentioned in the previous section, our approach limits the attack budget within 5\% of the total links in the graph to achieve the unnoticeable changes in the original graph. The visualizations demonstrate that the attacker applies the minor changes on the original graph, which can be considered as unnoticeable.

Upon observing Fig. \ref{visulization} (a) and \ref{visulization} (b), it is apparent that the two graph visualizations look very similar. Most of the communities are almost identical, indicating that the local structure of the graph is preserved. Since the ratio of modified links is much smaller than the total number of links, it is challenging to discern the detailed adversarial link modifications resulting from the proposed poisoned attack. These changes can be considered unnoticeable in the original graph. However, when observing Fig. \ref{visulization} (c) and \ref{visulization} (d), we notice that the overall appearance appears somewhat different. Most of the communities have changed, and even the sizes of the communities have changed. Upon closer inspection, Figs. \ref{visulization} (c) and \ref{visulization} (d) are more interconnected than Figs. \ref{visulization} (a) and \ref{visulization} (b). In addition, some new links have appeared in the upper right corner. To further investigate, we calculated the statistics of these graphs, as shown in Table \ref{statistic}. 

From Table \ref{statistic}, we observe that the statistics of Figs. \ref{visulization} (a) and \ref{visulization} (b) are very similar, only have slight differences in some aspects. The statistic of Fig. \ref{visulization} (c) and \ref{visulization} (d) have much more difference, especially on Fig. \ref{visulization} (d). From the visualization and statistical analysis, we draw several key observations. First, our adversarial modifications tend to add links to the original graph. Second, metrics such as average degree, average clustering coefficient, and average path length all show that the poisoned graph is more interconnected than the original. As the attack budget increases, this connectivity further intensifies.

\section{Conclusion}\label{sec:Conclusion}
In this article, we present an innovative unweighted graph poisoning attack that leverages meta-learning to degrade link-prediction performance. Our method employs a meta-learning framework with weighted schemes to identify optimal attack directions, enabling subtle yet effective modifications to the graph. Extensive experiments demonstrate the superior effectiveness of our attack and show the capability to identify the vulnerabilities of link prediction models. 


\bibliography{Reference}

\end{document}